\documentclass{article}

\usepackage{PRIMEarxiv}

\usepackage[utf8]{inputenc} 
\usepackage[T1]{fontenc}    
\usepackage{hyperref}       
\usepackage{url}            
\usepackage{booktabs}       
\usepackage{amsfonts}       
\usepackage{nicefrac}       
\usepackage{microtype}      
\usepackage{lipsum}
\usepackage{fancyhdr}       
\usepackage{graphicx}       
\usepackage{multicol}
\usepackage[utf8]{inputenc}
\usepackage{graphicx}
\usepackage{longtable}
\usepackage{geometry}
\usepackage{xtab}

\usepackage[
backend=biber,
style=ieee,
citestyle=numeric-comp
]{biblatex}

\AtEveryBibitem{\clearfield{address}}
\AtEveryBibitem{\clearlist{address}}

\AtEveryBibitem{\clearfield{place}}
\AtEveryBibitem{\clearlist{place}}

\AtEveryBibitem{\clearfield{issn}}
\AtEveryBibitem{\clearlist{issn}}

\AtEveryBibitem{\clearfield{series}}
\AtEveryBibitem{\clearlist{series}}

\AtEveryBibitem{\clearfield{month}}
\AtEveryBibitem{\clearlist{month}}

\AtEveryBibitem{\clearfield{isbn}}
\AtEveryBibitem{\clearlist{isbn}}

\AtEveryBibitem{\clearfield{language}}
\AtEveryBibitem{\clearlist{language}}

\AtEveryBibitem{\clearfield{doi}}
\AtEveryBibitem{\clearlist{doi}}

\graphicspath{{media/}}     

\author{
  Heinrich Peters\\
  Apple \\
  \texttt{\ heinrich\_peters2@apple.com} \\
   \And
  Alireza Hashemi\\
  Apple \\
  \texttt{\ arhash@apple.com} \\
   \And
  James Rae\\
  Apple \\
  \texttt{\ james\_r\_rae@apple.com} \\
  }

\title{Generalizable Error Modeling for Human Data Annotation: Evidence From an Industry-Scale Search Data Annotation Program}

\date{April 2021}

\begin{document}

\maketitle

\begin{abstract}
Machine learning (ML) and artificial intelligence (AI) systems rely heavily on human-annotated data for training and evaluation. A major challenge in this context is the occurrence of annotation errors, as their effects can degrade model performance. This paper presents a predictive error model trained to detect potential errors in search relevance annotation tasks for three industry-scale ML applications (music streaming, video streaming, and mobile apps). Drawing on real-world data from an extensive search relevance annotation program, we demonstrate that errors can be predicted with moderate model performance (AUC=0.65-0.75) and that model performance generalizes well across applications (i.e., a global, task-agnostic model performs on par with task-specific models). In contrast to past research, which has often focused on predicting annotation labels from task-specific features, our model is trained to predict errors directly from a combination of task features and behavioral features derived from the annotation process, in order to achieve a high degree of generalizability. We demonstrate the usefulness of the model in the context of auditing, where prioritizing tasks with high predicted error probabilities considerably increases the amount of corrected annotation errors (e.g., 40\% efficiency gains for the music streaming application). These results highlight that behavioral error detection models can yield considerable improvements in the efficiency and quality of data annotation processes. Our findings reveal critical insights into effective error management in the data annotation process, thereby contributing to the broader field of human-in-the-loop ML.
\end{abstract}

\keywords{Error Modeling \and Data Annotation \and Machine Learning \and Data Quality \and Human-in-the-loop ML}

\section{Introduction}
One of the key components of modern machine learning (ML) and artificial intelligence (AI) systems is the ever-increasing volume of human-annotated data used to train, evaluate, monitor, and update such models. The importance of this subject can hardly be overstated, as the quality of ML models is closely linked to the quality of the data they are trained and evaluated on \cite{sambasivan_everyone_2021, bernhardt_active_2022, amershi_modeltracker_2015, meek_characterization_2016, northcutt_pervasive_2021}. With the data annotation market reaching an estimated \$15 billion in 2023 \cite{verified_market_research_data_2023}, it is important to note that the impact of this industry extends far beyond this figure, as the revenue tied to the ML and AI products relying on human data annotation is much greater than the data annotation market itself. Therefore, a significant challenge lies in the management of \textbf{annotation errors} and their downstream consequences \cite{sambasivan_everyone_2021}.

In supervised ML, data annotation errors can impact model quality in at least two ways: First, mislabeling of training data is one of the main sources of prediction errors, as it prevents models from picking up on the true target function \cite{amershi_modeltracker_2015, meek_characterization_2016}. Second, mislabeling of testing data distorts model evaluations, therefore affecting model selection and leading to the deployment of suboptimal models \cite{northcutt_pervasive_2021}. For example, correcting labeling errors in the widely used ImageNet \cite{deng_imagenet_2009} benchmark dataset would change the rank order of model performance in favor of lower capacity models \cite{northcutt_pervasive_2021}.

Several potential solutions have been proposed to tackle annotation errors. These include developing clear annotation guidelines paired with extensive annotator training \cite{rottger_two_2022}, generating crowd-sourced annotations and monitoring inter-annotator agreement \cite{aroyo_truth_2015}, implementing audits, quality checks, or relabeling by subject matter experts \cite{bernhardt_active_2022}, as well as automated error detection techniques \cite{klie_annotation_2022, bernhardt_active_2022}. Generally, these practices should be implemented in a way that allows for feedback loops and iterative refinement. For example, detected errors can be used as an opportunity to refine guidelines and provide additional training to annotators. Similarly, annotators should be able to ask questions or raise concerns about the annotation process or specific items to help identify problems early \cite{tseng_best_2020, mcdonnell_why_2016}.

The present paper explores automated error models and their potential to improve the efficiency of the data annotation process, specifically with respect to the quality and turnaround of annotations. We further analyze how automated error detection models can improve related processes such as auditing and relabeling. Compared to previous work addressing annotation errors, our approach introduces a unique emphasis on task-agnostic error prediction. Unlike many prior models, which focus on specific tasks and require separate training for each labeling task, our models are trained to predict errors directly from a combination of task features and behavioral features derived from the annotation process itself, offering a higher degree of generalizability.

\subsection{Background and Related Work}
Automated error detection techniques aim to identify instances that may be labeled incorrectly or inconsistently within a given dataset \cite{klie_annotation_2022}. Once these potential errors are detected, they can be forwarded for manual review by human annotators or processed via other annotation error correction methods. Error detection has been implemented across diverse domains in natural language processing \cite{barnes_sentiment_2019, alt_tacred_2020, klie_annotation_2022}, and image classification \cite{northcutt_pervasive_2021, bernhardt_active_2022}. However, the logic of error modeling can be applied to a variety of supervised ML tasks.

Previous literature has distinguished between variation-based, rule-based, or model-based approaches to error detection \cite{klie_annotation_2022}. Variation-based approaches generally rely on inputs labeled by multiple annotators. For example, in settings where only one label can be correct, disagreement across annotators can indicate annotation errors. Annotator disagreement can be captured by computing variation metrics like label entropy across annotators, such that high entropy would be indicative of annotation errors \cite{hollenstein_inconsistency_2016}. While variation-based error detection methods are easy to implement, one major disadvantage is that they are limited to crowd-sourced scenarios, where more than one annotator label is produced for each instance. Requiring multiple labels can introduce significant costs to annotation processes.

Rule-based techniques, on the other hand, are useful when there are clear annotation guidelines that can be translated into formal rules. In such situations, annotations that are impossible given the set of rules would be flagged as incorrect. For example, errors can be flagged if an annotator picks a label that is not applicable to the given task. In language annotation tasks, n-grams that are lexically or morphologically impossible would fall into this category \cite{kveton_semi-automatic_2002}. However, while rule-based approaches can prove to be highly effective, they are typically tailored to a very specific task, imposing limits on generalizability. Additionally, they only apply to a specific subset of errors, namely those that are the result of objective rule violations.

Model-based error detection approaches typically utilize probabilistic classifiers designed to detect annotation errors. Past work has frequently relied on the models predicting an annotation label (typically a multi-class classification problem) and subsequently comparing the predicted label to a label provided by a human annotator. If there is disagreement between the human annotation and the model-generated annotation, the annotation can be marked as inconsistent and additionally directed to a human auditor \cite{larson_inconsistencies_2020, van_halteren_detection_2000}. Alternatively, model confidence scores or predicted probabilities can provide an indication of task difficulty \cite{amiri_spotting_2018}. Implicitly, these approaches rely on the assumption that models perform worse on instances that are more difficult for human annotators as well. Consequently, instances that show low model performance - either by deviation from human annotations or by low model confidence scores - would be more likely to contain annotation errors.

However, it is important to note that a model's confidence score on the labeling task itself may not always be a reliable indicator of a human annotator's potential error. For instance, low model confidence could be due to the model's limitations in representing complex or context-specific tasks rather than indicating an annotation error. Conversely, there may be cases where a model is highly confident and yet wrong due to biases in training data or overly simplistic decision boundaries. Therefore, while using model confidence scores can certainly be a useful tool in identifying potential annotation errors, it should be treated as part of a broader toolkit and used in conjunction with other methods. For a more comprehensive approach to error detection, it might be beneficial also to consider other factors such as the context of the task, the difficulty level of the task, and the reliability and expertise of the human annotators themselves. Relatedly, model-based approaches can be less effective in identifying systematic errors, where both the model and the human annotators consistently misinterpret certain types of instances. Such errors could go undetected as model predictions and human annotations would agree, despite being incorrect. Finally, approaches predicting labels may suffer from low generalizability, as a separate model needs to be trained for each specific labeling task, which can be computationally expensive and may not be feasible for all tasks, especially those requiring complex models.

As an alternative approach explored in this paper, a model can be used to directly predict whether a human label is correct or not (typically a binary classification problem) without explicitly predicting the underlying label itself. The advantage of this approach is that it is often easier to predict the presence of an error compared to predicting the correct label. Relatedly, directly predicting the correctness of a label can be computationally more efficient, as it does not necessarily need to consider all possible label classes, which can be vast in certain tasks. Overall this approach makes it easier to train, evaluate, and maintain such models even when training data are limited. Additionally, error detection models can be set up to be task agnostic: a model predicting errors instead of labels may generalize to a wider range of tasks because certain features are applicable to many different scenarios. For example, taking into account general measures of task complexity, past annotator performance, time on task, or comments provided by annotators can be informative irrespective of the underlying labeling task. Past research has rarely applied this approach \cite{klie_annotation_2022} since most studies are based on archival data for which no annotator data (e.g., past performance) or task-completion data (e.g., time on task) are available. 

\subsection{Current Research}

The current research presents a study on generalizable error detection models, taking into account behavioral features in addition to task features, and their applications for search relevance annotation tasks. The tasks were drawn from an extensive data annotation program that generates training and evaluation data for industry-scale search and recommendation systems. We focus on search relevance annotations related to music streaming, video streaming, and mobile applications. A detailed description of the annotation task can be found in the Method section below.

Drawing on authentic data from the search relevance annotation program, we first use model-based error detection techniques to show that annotation errors can be detected with moderate classification performance, based on a combination of task features and behavioral features derived from the data annotation process. We demonstrate the predictability of errors in search relevance rating tasks for all three ML applications: music streaming, video streaming, and mobile applications. Second, using Shapley values \cite{lundberg_unified_2017}, we explore the contributions of a variety of behavioral and task-related features in order to better understand which sources of information are driving the model’s predictions. Third, we explore the generalizability of our error detection models across rating tasks and ML applications. Fourth, we show that the error detection models can unlock considerable efficiency gains for auditing and relabeling by prioritizing tasks with high predicted error probabilities at the expense of tasks with low predicted error probabilities. Taken together, our findings show that behavioral error detection models can be highly effective in improving the efficiency of data annotation processes.

\section{Method}
\subsection{Data Collection and Sampling}
Our study is based on authentic data from an industry-scale search relevance annotation program. We used a large archival dataset of audited search relevance annotation tasks, including metadata, focused on music streaming, video streaming, and mobile applications. Audited tasks include annotator labels as well as ground truth labels from subject matter experts who were involved in the development of the annotation guidelines. We restricted the data to audited tasks in order to calculate error rates against ground truth labels provided by the subject matter experts. The dataset contained search relevance annotations for 42,611 individual tasks from 751 annotators spanning 23 storefronts (countries). From this dataset, 30\% of observations were randomly selected for a holdout test set.

\subsection{Search Relevance Annotation Tasks}
The search relevance annotation tasks we used were presented as query-output pairs for which a human annotator provided a rating in one of 5 categories: Perfect, Excellent, Good, Acceptable, or Unacceptable. Each of the label categories was defined according to a set of rules and examples specified in comprehensive, task-specific guidelines provided to the annotators. In the case of music streaming, for example, returning the eponymous Metallica song for the query “Master of Puppets” would be rated as “Perfect”, while a Taylor Swift song would be rated as “Unacceptable”.  In cases where none of the labels would apply, annotators could flag a problem in the annotation task. The current study covers a total of three different search relevance rating tasks for music streaming, video streaming, and mobile applications. Importantly, while each query would typically prompt a range of outputs or suggestions in each product’s user interface (UI), for the purpose of search relevance annotations, the individual outputs were rated in isolation. In other words, annotators only saw one query-output pair at a time and were blind to the ranking of the respective output item in the overall list of search results. Tasks were audited by subject matter experts who could overturn the original annotator labels.

\begin{table*}[htbp]
\footnotesize

\renewcommand{\arraystretch}{1.2}  
\centering
\begin{xtabular*}{\textwidth}{p{6.8cm}p{9cm}}
\hline
\textbf{Feature Name} & \textbf{Description} \\
\hline
\multicolumn{2}{l}{\textit{Task Features}} \\ 
\cmidrule(lr){1-2}
\hspace{1em}input\_occurrences & Cumulative number of searches for focal query (reflects popularity) \\
\hspace{1em}input\_conversion\_rate & Proportion of users who converted on any output \\
\hspace{1em}in\_out\_spacy\_distance & Cosine similarity between word embeddings of input and output string \\
\hspace{1em}in\_out\_edit\_distance & Edit distance between input and output string \\
\hspace{1em}input\_query\_type & Intent classification for query \\
\hspace{1em}input\_media\_type & Indicates whether query was made through keyboard or verbally through a voice assistant \\
\hspace{1em}output\_media\_type & Type of search results in six different content categories \\
\hspace{1em}input\_misspelled & Binary variable indicating if the input was misspelled \\
\hspace{1em}input\_language & Categorical variable distinguishing between 21 input languages \\
\hspace{1em}storefront\_name & Categorical variable distinguishing between 23 geographic regions \\
\cmidrule(lr){1-2}
\multicolumn{2}{l}{\textit{Past Performance Features}} \\ 
\cmidrule(lr){1-2}
\hspace{1em}error\_{N} & Rolling window of N = [7, 14, 21, 28] days for error rate of focal annotator (4 features) \\
\hspace{1em}error\_N\_all & Rolling window of N = [7, 14, 21, 28] days for error rate of all annotators in same eval (4 features) \\
\hspace{1em}maj\_error\_N & Rolling window of N = [7, 14, 21, 28] days for major error (annotator label more than 2 levels off) rate of focal annotator (4 features) \\
\hspace{1em}maj\_error\_N\_all & Rolling window of N = [7, 14, 21, 28] days for major error (annotator label more than 2 levels off) rate for all annotators in same eval (4 features) \\
\hspace{1em}error\_N\_diff & Difference between annotator error rate and overall error rate of N = [7, 14, 21, 28] day rolling windows (4 features) \\
\hspace{1em}vol\_last\_N & Rolling window of N = [7, 14, 21, 28] days for task volume of focal annotator \\
\hspace{1em}tenure\_full\_days & Number of days a rater has been active since last deactivation \\
\hspace{1em}tenure\_updated\_days & Number of days a rater has been active since first joining the rater pool \\
\hspace{1em}qualification\_trials & Number of trials an annotator took to pass the qualification test \\
\hspace{1em}qualification\_agreement\_rate & Performance on qualification test \\
\hspace{1em}error\_rolling\_output\_media\_type\_N & Share of correct annotations by focal annotator for given media type over window of N = [1, 3, 5] (3 features) \\
\hspace{1em}error\_rolling\_input\_query\_type\_user\_N & Share of correct annotations by focal annotator for given input query type over window of N = [1, 3, 5] (3 features) \\

\cmidrule(lr){1-2}
\multicolumn{2}{l}{\textit{Session Context Features}} \\ 
\cmidrule(lr){1-2}
\hspace{1em}nth\_task\_in\_session & Counter indicating the position of the current task within an ongoing session \\
\hspace{1em}seconds\_into\_session & Time that has passed since the start of the session in seconds \\
\cmidrule(lr){1-2}
\multicolumn{2}{l}{\textit{Task-Completion Features}} \\ 
\cmidrule(lr){1-2}
\hspace{1em}answer\_value & Label provided by annotator \\
\hspace{1em}time\_on\_task & Time (in seconds) spent on focal annotation task \\
\hspace{1em}comment\_length & Number of words in annotator comment \\
\hline
\end{xtabular*}
\caption{Overview of features used in error models. Please refer to this table when interpreting the feature importance scores in Figure 2.}
\label{tab:feature_table}
\end{table*}

\subsection{Operationalizations and Data Preprocessing}
The features used in the models fall into four broad categories: Task metadata (e.g., input-query strings, search-output strings, historical conversion rates, etc.), past annotator performance (e.g., moving averages of performance or task volume over various time windows), session context features (e.g., task position in session), and task-completion features representing an annotator's interaction with a specific task (e.g., time on task, labels provided by the annotator, etc.). The latter three groups are behavioral features derived from the data annotation process. For a full list of features, please refer to Table \ref{tab:feature_table}.

We used standard preprocessing techniques. Mean imputation and standard scaling were used for numerical features. For categorical features, missing values were imputed with a unique “missing” category before applying one-hot encoding. All transformations were performed relative to the distribution of values in the training set, such that no information was leaked from the validation and test set.

\subsection{Modeling}
The error detection problem can be cast as a binary classification task, where the positive class denotes annotation errors. An error is defined as an annotation that violates annotation guidelines, as established by an expert auditor. To tackle this classification task, we used XGBoost \cite{chen_xgboost_2016}, an ML algorithm known for its suitability across a diverse range of prediction tasks. XGBoost employs gradient boosting, an ensemble technique that builds a predictive model through the sequential combination of weak learners, typically decision trees. The process begins with a single tree making an initial prediction. In subsequent iterations, new trees are created to correct the residuals from the previous trees, progressively refining the model. The final model is a weighted sum of the trees generated during the iterative training process. This approach generally achieves high predictive performance on structured data, often surpassing other state-of-the-art ML techniques \cite{shwartz-ziv_tabular_2021}.

In line with standard procedures for hyperparameter tuning, we performed randomized search over a comprehensive hyperparameter space (see SI A). Model performance for different hyperparameter configurations was assessed on a validation set that had not been previously used for training. The validation set was split off the training set such that 30\% of the data points were used for validation. We chose AUC as the selection criterion in order to account for the imbalanced nature of the target variable. The hyperparameter configuration resulting in the highest AUC on the validation set was used to retrain the model on the union of the training and validation sets. This final model was then evaluated on the test set. We report standard evaluation metrics for binary classification (AUC, accuracy, precision, and recall), based on a predicted probability score cutoff of 0.5.

We trained four distinct error models: One specialized model for each of the three ML applications (music streaming, video streaming, and mobile applications), as well as one global, task-agnostic model trained to perform well in all three of these settings. The specialized models were trained, selected, and evaluated on application-specific subsets of the data (e.g., only music streaming), while the task-agnostic model was trained on the union of all datasets.

\section{Results}
\subsection{Model Performance}
\begin{figure*}[htbp]
    \centering
    \includegraphics[width=0.9\textwidth]{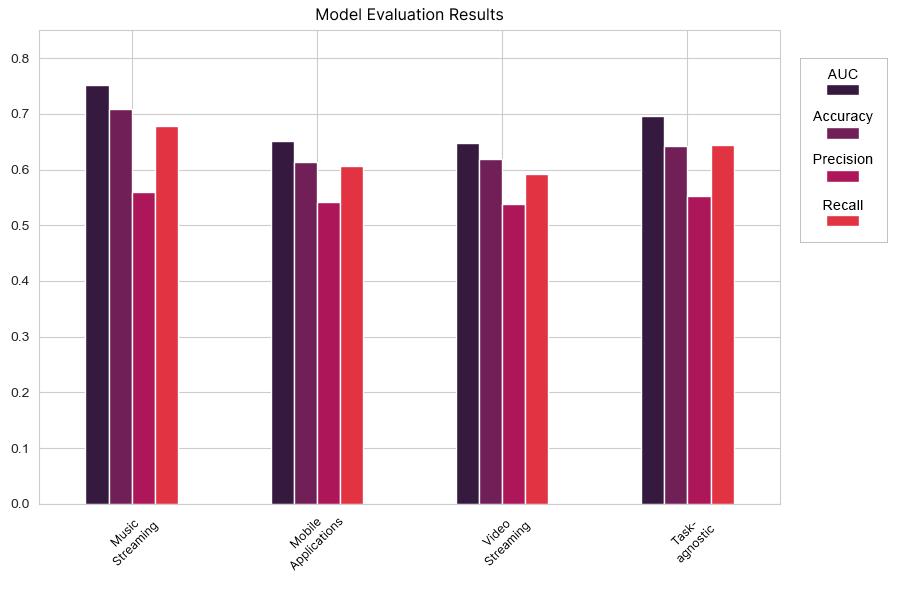}
    \caption{Area under the receiver operating characteristics curve (AUC), accuracy, precision, and recall of error models trained for different ML applications and a task-agnostic model trained for all ML applications simultaneously. Precision and recall were computed as macro average across both classes. For a table showing exact numbers, please refer to SI B.}
    \label{fig:model_eval_results}
\end{figure*}

Model performance was evaluated on a test set of 14,202 (music streaming = 6,151; mobile applications = 6,075; video streaming = 1,976) audited annotation tasks. In all evaluations, annotation errors were assigned the positive class label. Overall classification performance was moderate in all models. The model for music streaming performed best (AUC=0.75, accuracy=0.71, precision=0.56, recall=0.68), followed by mobile applications (AUC=0.65, accuracy=0.61, precision=0.54, recall=0.61), and video streaming (AUC=0.65, accuracy=0.62, precision=0.54, recall=0.59). The task-agnostic model showed comparable performance on the combined test sets (AUC=0.70, accuracy=0.64, precision=0.55, recall=0.64). For a graphical overview of model performance please refer to Figure \ref{fig:model_eval_results}. Exact values can be found in SI B.

\subsection{Model Explainability}
\begin{figure*}[htbp]
    \centering
    \includegraphics[width=\textwidth]{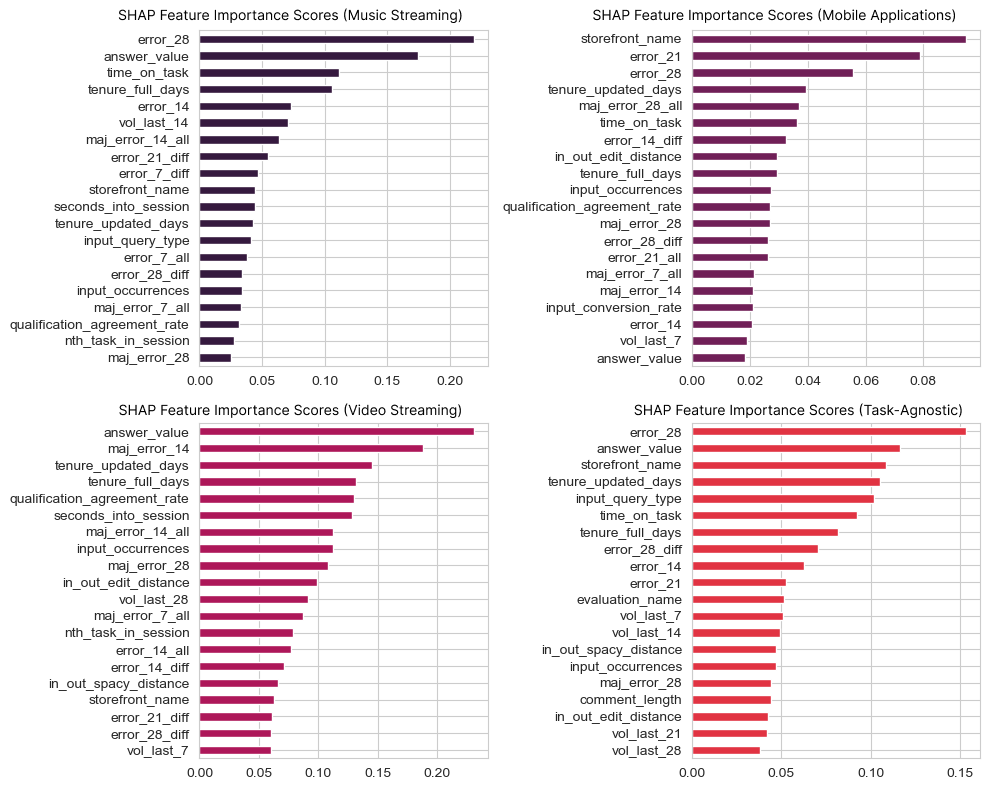}
    \caption{Cumulative mean SHAP values, ranked by magnitude, for error models trained for different ML applications and a task-agnostic model trained for all ML applications simultaneously. For a table showing exact numbers, please refer to SI E.}
    \label{fig:feature_importance}
\end{figure*}

In order to gain deeper insights into the feature-target associations driving model performance, we generated SHAP values \cite{lundberg_unified_2017} for all models (see Figure \ref{fig:feature_importance}). We used the training set for each model as a background set and then computed the SHAP values on the respective test set. In all four models, features related to past annotator performance (e.g., error\_28, error\_14) as well as experience (e.g., tenure\_days\_updated, tenure\_full\_days, vol\_last\_21, etc.) were among the most predictive. Our models also strongly relied on task-related features (e.g., storefront\_name, input\_occurrences, in\_out\_edit\_distance) and task-completion features (answer\_value, time\_on\_task).

In summary, while the rank order of features was somewhat different from model to model, all models were driven by a mix of features related to past annotator performance, task-related and session-context features, and task-completion features.

\subsection{Generalizability}
We purposefully chose to construct a feature set that was expected to generalize across a wide range of search relevance tasks associated with vastly different ML applications. In order to test the generalizability of our error models, we applied the models trained to detect errors for a specific ML application to other ML applications and assessed their performance on this new data. For example, we would evaluate a model trained on music streaming search results on a test set of mobile applications search results. We performed this validation procedure for all combinations. If performance degrades strongly when moving from one application to another, this might indicate that some features are very specific to a particular application and may not generalize well. 

\begin{figure*}[tbp]
    \centering
    \includegraphics[width=\textwidth]{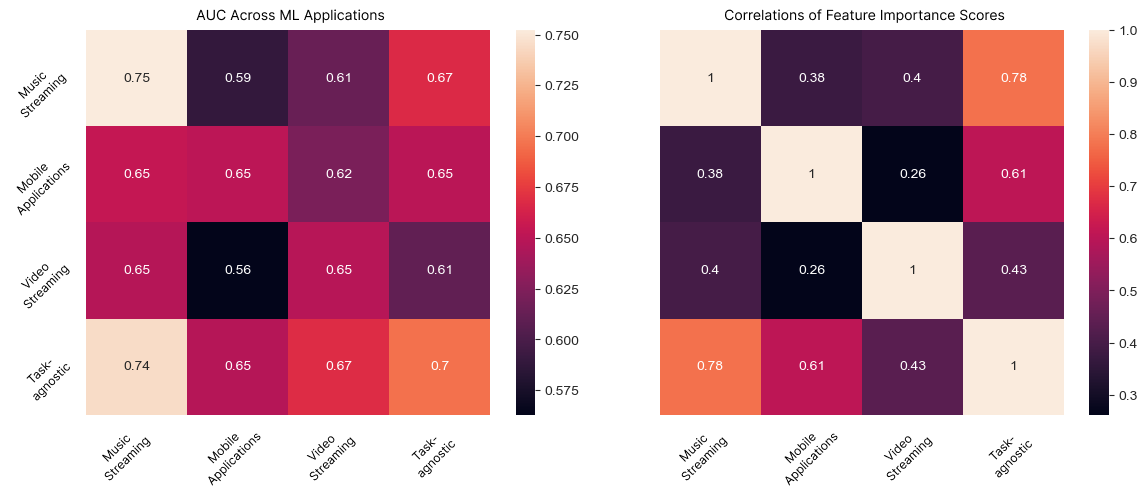}
    \caption{AUC matrix of generalization performance across search modalities and product categories (left; the training set is denoted on the y-axis; test set is denoted on the x-axis). Correlations of SHAP feature importances across search modalities and product categories (right). Exact values can be found in SI C and D.}
    \label{fig:generalizability}
\end{figure*}

Our results show that the models generally performed considerably better than chance, even across ML applications (music streaming, mobile applications, video streaming). As expected, generalized model performance of task-specific models was somewhat lower on test sets for other ML applications. Importantly, however, the task-agnostic model showed satisfactory performance on all three ML applications and performed on par with the specialized models, even on their native test sets (AUC = 0.74 vs AUC = 0.75 for music streaming, AUC = 0.65 vs AUC = 0.65 for mobile applications, AUC = 0.67 vs AUC = 0.65 for video streaming; see Figure \ref{fig:generalizability}).

Furthermore, we explored whether feature importance scores generalized across ML applications. For this purpose, we computed the correlation between mean-absolute SHAP values (at feature level) across ML applications. The results generally show positive relationships, indicating that the same features tend to be informative across ML applications. However, the magnitude of correlations varied somewhat. The task-agnostic model shows the highest correlations across the board, reflecting the fact that the model was partly trained on the same data as specialized models.

Taken together, our findings show that error models generalize well across ML applications. This indicates that the underlying structure of the error detection problem may be similar across different search relevance tasks, given the chosen feature set of task-related and behavioral features. In practice, this means that teams that have multiple annotation programs can maintain a single error model for a variety of annotation tasks rather than relying on many different task-specific models, which would be harder to develop and maintain.

\subsection{Model Application: Error Detection for Efficient Relabeling}
As explained in the introduction, auditing and relabeling are an integral but expensive aspect of data annotation processes. Given that audits and relabeling are expensive, it is efficient to focus these efforts on instances where the likelihood of finding incorrect labels is the highest. In order to estimate the efficiency gains for relabeling provided by the error detection model, we compared a relabeling schedule that was based on predicted error probabilities to a random sampling strategy. Specifically, we ranked annotation tasks by their predicted error probabilities and analyzed the proportion of labels flipped during auditing (i.e., changed from an incorrect label to the correct one), relative to the number of audited tasks, and relative to error volume.

\begin{figure*}[tbp]
    \centering
    \includegraphics[width=\textwidth]{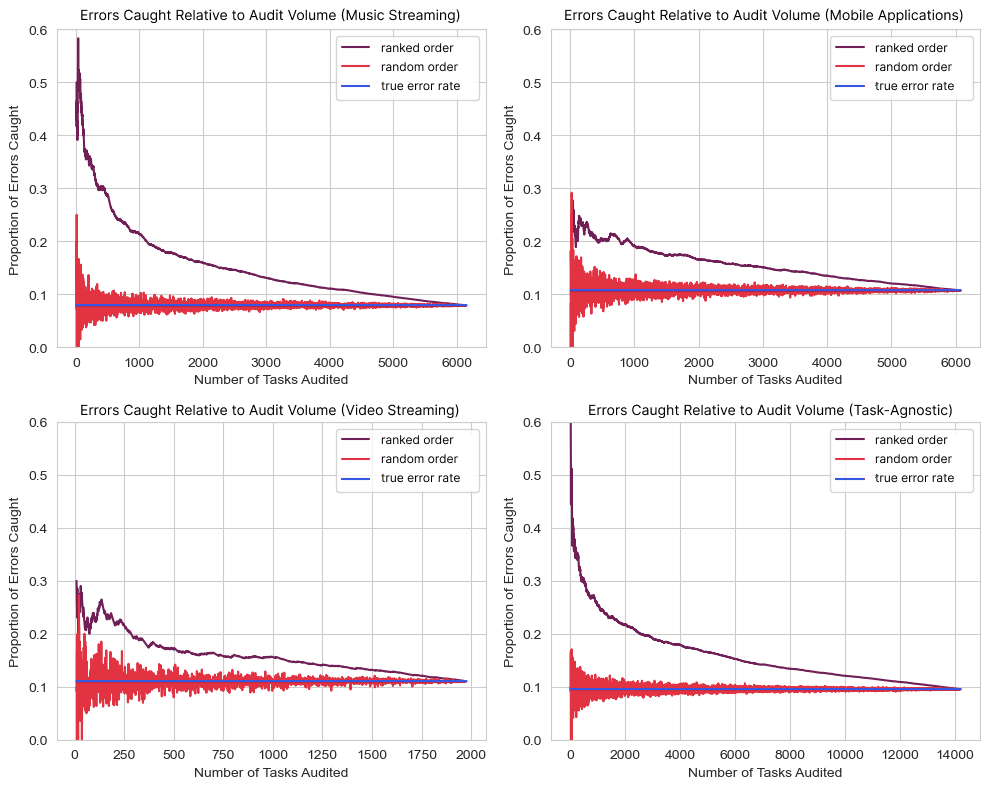}
    \caption{Proportion of labels changed relative to audit volume as a function of the number of audited tasks for music streaming, mobile applications, video streaming, and all three ML applications combined.}
    \label{fig:audit_1}
\end{figure*}

First, we analyzed the proportion of labels flipped relative to the number of audited tasks (Figure \ref{fig:audit_1}). This analysis demonstrates that prioritizing tasks with high predicted error probabilities considerably increases the number of corrected errors early in the audit process. While the expectation of flipping an error using a random sampling strategy is equal to the overall error rate (around 10\% in our case), the expectation of flipping a label for tasks with high predicted error probabilities is considerably higher. For example, the first 100 tasks audited for music streaming would show a more than 4-times increased likelihood of catching an error for each audited task. This benefit levels off as more tasks are audited, but efficiency gains can be realized even for relatively high audit volumes.

\begin{figure*}[tbp]
    \centering
    \includegraphics[width=\textwidth]{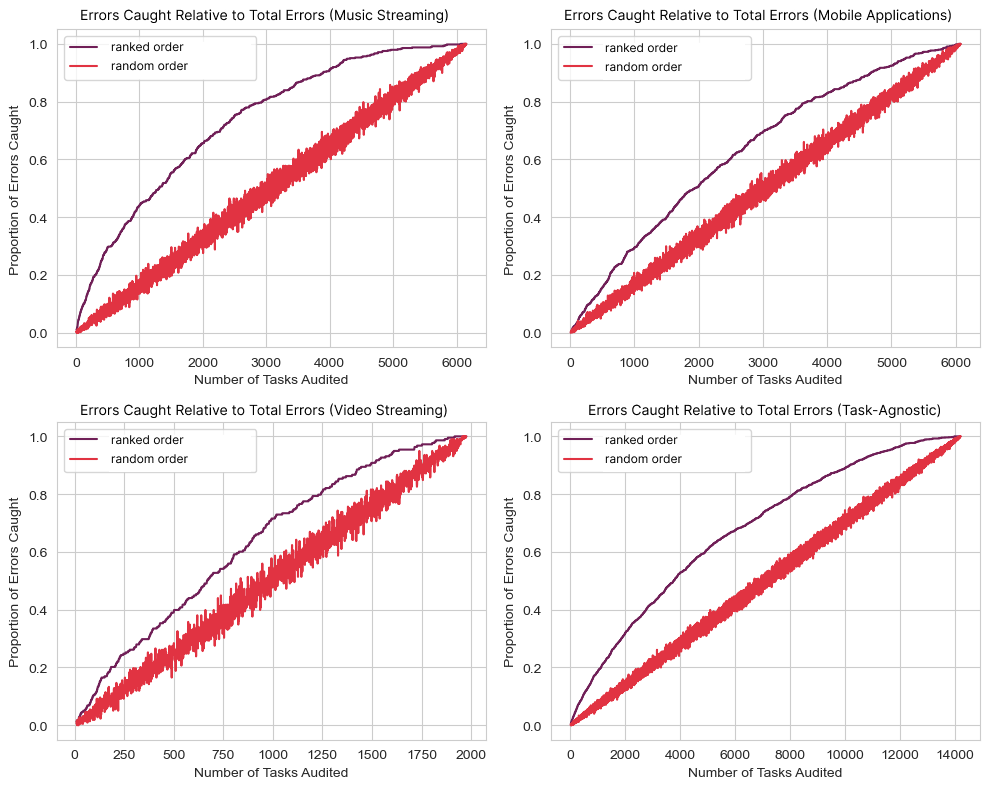}
    \caption{Proportion of errors caught relative to overall number of errors as a function of the number of audited tasks for music streaming, mobile applications, video streaming, and all three ML applications combined.}
    \label{fig:audit_2}
\end{figure*}

A similar but distinct perspective is shown in Figure \ref{fig:audit_2}, where we analyze the number of errors caught as a proportion of the overall number of errors present in the set of annotation tasks. The diagonal shows the proportion of errors caught as a function of the number of audits with random sampling. Here, we would need to audit roughly X\% of the tasks in order to catch X\% of annotation errors. For example, if we wanted to catch 80\% of errors we would have to audit 80\% of the tasks with random sampling. Using model-based ranking, however, increases the proportion of errors caught early on to the extent that the curved purple line deviates from the red diagonal, which is a direct reflection of model performance. In the case of music streaming, for example, this means we would have to audit roughly 5,000 tasks with random sampling but only roughly 3,000 tasks with model-based task ranking in order to catch 80\% of errors. This efficiency gain of 40\% would translate into an equivalent reduction in audit costs while achieving the same results. The depiction in Figure \ref{fig:audit_2} presents an easy and intuitive approach for estimating the number of audits that are needed in order to relabel a target proportion of annotation errors.

Our analyses indicate considerable efficiency gains for auditing and relabeling when tasks are selected based on model-generated error probabilities as compared to random sampling, even for models with moderate classification performance.

\section{Discussion}
\subsection{Interpretation of Results}
Our study evaluates the performance of models predicting annotation errors on a test set of audited search relevance rating tasks for music streaming, mobile applications, and video streaming.
Model performance was considerably better than chance across all use cases. The drivers of model performance include past annotator performance and experience, task-related features, session context, and task-completion features. Models generalized relatively well over different ML applications. The task-agnostic model performed on par with task-specific models across annotation tasks for all three ML applications, indicating that task-specific models can potentially be replaced by task-agnostic models. Finally, model-based error predictions were found to provide considerable efficiency gains for relabeling and auditing tasks when compared to a random sampling strategy. Overall, these findings underline the utility of behaviorally informed error detection models across a wide range of search relevance rating tasks.

Our work ties in with the past literature on error modeling as it demonstrates the feasibility of error modeling in human annotation tasks. However, our approach and results also deviate from prior work in several important ways. First, our approach presents a shift in methodology: Instead of first predicting labels and then analyzing disagreement with human-generated annotations \cite{klie_annotation_2022, van_halteren_detection_2000, larson_inconsistencies_2020}, we model errors directly from a mix of task features, annotator features, session context features, and task-completion features. Second, while previous efforts predominantly focused on applications in natural language processing \cite{barnes_sentiment_2019, alt_tacred_2020, klie_annotation_2022} and image classification \cite{northcutt_pervasive_2021, bernhardt_active_2022}, we demonstrate that error modeling can be successfully applied to search relevance rating tasks. The high efficacy of behaviorally informed features is aligned with previous work employing psychologically motivated human error frameworks \cite{pandey_modeling_2022}. With regard to auditing and relabeling, past work has mostly suggested disagreement-based sampling to select tasks. In a multiple coverage scenario, tasks with significant disagreement among the annotators are prime candidates for auditing or relabeling. In single-coverage scenarios, disagreement between a human annotator and a model trained to predict labels \cite{van_halteren_detection_2000, larson_inconsistencies_2020, amiri_spotting_2018} can indicate the need for relabeling. Our work is aligned with this approach insofar as it suggests a model-based approach to select instances for relabeling. However, directly modeling annotation errors is arguably simpler, more efficient, and more generalizable. Moreover, our results are aligned with previous findings that ranking tasks by estimated label correctness and labeling difficulty can dramatically improve relabeling efficacy in image processing \cite{bernhardt_active_2022}.

\subsection{Implications, Limitations, and Directions for Future Work}
The development and successful implementation of error prediction models can enhance the efficiency of data annotation processes in practice. For relabeling and auditing, the usage of these models can lead to significant time and cost savings (e.g., 40\% in the case of music streaming) by prioritizing tasks based on model-generated error probabilities. The proposed method would improve the overall quality of training and evaluation data, making it a more affordable and reliable input for ML systems. With the continuing rise in demand for human-annotated data, for example in the context of training large language models (LLMs) \cite{ouyang_training_2022}, the use of error models and other helper systems will only continue to grow in relevance. 

However, it is important to consider that the use of predictive error models may lead to unintended consequences. First, in the context of relabeling, auditors’ ratings may be biased by their knowledge of error probabilities. For example, auditors may be overly prejudiced against prioritized audits if they are aware of the fact that the most suspicious labels are always presented first. To address this potential bias in relabeling, a blind task allocation process or intentional opacity around error probabilities should be maintained. For example, when a predefined subset of tasks needs to be audited, they can be chosen in a model-based fashion, but the selected tasks should be presented in random order. Importantly, a model-based auditing strategy, while beneficial for relabeling, might make it more challenging to estimate error rates effectively. This is because error rates will exhibit a strong upward bias when tasks are selected based on predicted error probabilities. Effectively correcting for this type of sampling bias should be a goal of future research. Finally, there is also the risk of model drift \cite{gama_learning_2004}: As the allocation of annotation tasks changes based on the predictions of the model, so too does the distribution of task features and error rates, which could cause model performance to deteriorate over time. To combat model drift, it is important to implement regular monitoring and frequent model updates.

Another practical challenge is the integration of predictive error models into existing data annotation platforms. Conventional approaches that focus on label prediction do not necessitate the computation of additional features, such as past annotator performance, task-completion, and session features. While the integration of such behavioral features may lead to generalizable models, the additional data collection can be resource-intensive and may require the construction of sophisticated data processing pipelines in production settings. The benefits of potential improvements in error detection must be weighed against these costs, likely favoring the proposed behavioral modeling approach in scenarios with high annotation volume.

Future work should aim to further enhance the accuracy and generalizability of error detection models. It would be particularly interesting to explore the application of these models to other types of annotation tasks beyond search relevance ratings. This could include the training of LLMs, where annotation guidelines may be less well-defined and harder to enforce due to the inherent subtlety of natural language. Importantly, the rise of LLMs also foreshadows a shift in data annotation practices, as LLMs do not only rely on annotated data for training but may become helpful in data annotation tasks and error modeling tasks themselves. 

Furthermore, our model could contribute to different applications such as label aggregation \cite{zheng_truth_2017} or task routing \cite{yang_predicting_2019}, which were beyond the scope of the current paper. Label aggregation may be necessary in scenarios where multiple annotators have generated a label for the same input/task to derive a single annotation based on the annotations of multiple individual annotators \cite{zheng_truth_2017}. Annotation errors can hinder the aggregation process by adding bias or noise. Using an error model to exclude annotations likely containing errors may therefore make the label aggregation process more efficient. Task routing, on the other hand, refers to the process of assigning specific tasks to appropriate workers based on their skills, expertise, or past performance \cite{yang_predicting_2019}. Error models such as ours could be helpful in assigning tasks such that the overall likelihood of annotation errors is minimized. One requirement for routing is that the model should be capable of predicting errors before the annotation task has been assigned. Such a model, therefore, could not contain task-completion features. Interestingly, many of the features used in our current models are available pre-annotation (e.g., task features, past annotator performance, session context features), making it possible to train prospective error models, for which we have obtained encouraging results in a series of pilot studies. The use of prospective error models for intelligent task routing will be investigated in future work. 

\subsection{Conclusion}

Our findings demonstrate the potential of behavioral error detection models to enhance the efficiency and quality of large-scale data annotation processes. By directly modeling errors using a mix of task features, annotator features, session-context features, and task-completion features, we were able to achieve moderate predictive performance on a range of search relevance rating tasks. The models generalized well across applications and were shown to provide dramatic efficiency gains for relabeling and auditing. As ML and AI systems are increasingly deployed across various industries, ensuring high-quality training and evaluation data is imperative. Our work shows that data quality can be improved in a cost-effective manner using behaviorally informed predictive models.


\section*{Author Contributions}
HP: Conceptualization, Data Curation, Formal Analysis, Investigation, Methodology, Software, Visualization, Writing - Original Draft; AH: Conceptualization, Data Curation, Methodology, Software, Supervision,  Writing – Review \& Editing; JR: Conceptualization, Methodology, Supervision, Project Administration, Writing – Review \& Editing

\section*{Acknowledgements}
We thank Nilab Hessabi, Sanjay Srivastava, Samir Patel, and Emily Minette for their support.

\newpage
\printbibliography

@inproceedings{lundberg_unified_2017,
	address = {Red Hook, NY},
	series = {{NIPS}'17},
	title = {A unified approach to interpreting model predictions},
	isbn = {978-1-5108-6096-4},
	urldate = {2021-12-22},
	booktitle = {Proceedings of the 31st {International} {Conference} on {Neural} {Information} {Processing} {Systems}},
	publisher = {Curran Associates Inc.},
	author = {Lundberg, Scott M. and Lee, Su-In},
	month = dec,
	year = {2017},
	pages = {4768--4777},
}

@article{aroyo_truth_2015,
	title = {Truth {Is} a {Lie}: {Crowd} {Truth} and the {Seven} {Myths} of {Human} {Annotation}},
	volume = {36},
	copyright = {Copyright (c)},
	issn = {2371-9621},
	shorttitle = {Truth {Is} a {Lie}},
	url = {https://ojs.aaai.org/aimagazine/index.php/aimagazine/article/view/2564},
	doi = {10.1609/aimag.v36i1.2564},
	language = {en},
	number = {1},
	urldate = {2023-05-15},
	journal = {AI Magazine},
	author = {Aroyo, Lora and Welty, Chris},
	month = mar,
	year = {2015},
	note = {Number: 1},
	pages = {15--24},
}

@article{zheng_truth_2017,
	title = {Truth inference in crowdsourcing: is the problem solved?},
	volume = {10},
	issn = {2150-8097},
	shorttitle = {Truth inference in crowdsourcing},
	url = {https://dl.acm.org/doi/10.14778/3055540.3055547},
	doi = {10.14778/3055540.3055547},
	number = {5},
	urldate = {2023-07-19},
	journal = {Proceedings of the VLDB Endowment},
	author = {Zheng, Yudian and Li, Guoliang and Li, Yuanbing and Shan, Caihua and Cheng, Reynold},
	month = jan,
	year = {2017},
	pages = {541--552},
}

@inproceedings{yang_predicting_2019,
	address = {Minneapolis, Minnesota},
	title = {Predicting {Annotation} {Difficulty} to {Improve} {Task} {Routing} and {Model} {Performance} for {Biomedical} {Information} {Extraction}},
	url = {http://aclweb.org/anthology/N19-1150},
	doi = {10.18653/v1/N19-1150},
	language = {en},
	urldate = {2023-07-13},
	booktitle = {Proceedings of the 2019 {Conference} of the {North}},
	publisher = {Association for Computational Linguistics},
	author = {Yang, Yinfei and Agarwal, Oshin and Tar, Chris and Wallace, Byron C. and Nenkova, Ani},
	year = {2019},
	pages = {1471--1480},
}

@inproceedings{larson_inconsistencies_2020,
	address = {Barcelona, Spain (Online)},
	title = {Inconsistencies in {Crowdsourced} {Slot}-{Filling} {Annotations}: {A} {Typology} and {Identification} {Methods}},
	shorttitle = {Inconsistencies in {Crowdsourced} {Slot}-{Filling} {Annotations}},
	url = {https://www.aclweb.org/anthology/2020.coling-main.442},
	doi = {10.18653/v1/2020.coling-main.442},
	language = {en},
	urldate = {2023-06-30},
	booktitle = {Proceedings of the 28th {International} {Conference} on {Computational} {Linguistics}},
	publisher = {International Committee on Computational Linguistics},
	author = {Larson, Stefan and Cheung, Adrian and Mahendran, Anish and Leach, Kevin and Kummerfeld, Jonathan K.},
	year = {2020},
	pages = {5035--5046},
}

@inproceedings{kveton_semi-automatic_2002,
	address = {Taipei, Taiwan},
	title = {({Semi}-)automatic detection of errors in {PoS}-tagged corpora},
	volume = {1},
	url = {http://portal.acm.org/citation.cfm?doid=1072228.1072249},
	doi = {10.3115/1072228.1072249},
	language = {en},
	urldate = {2023-06-30},
	booktitle = {Proceedings of the 19th international conference on {Computational} linguistics  -},
	publisher = {Association for Computational Linguistics},
	author = {Květoň, Pavel and Oliva, Karel},
	year = {2002},
	pages = {1--7},
}

@inproceedings{amershi_modeltracker_2015,
	address = {Seoul Republic of Korea},
	title = {{ModelTracker}: {Redesigning} {Performance} {Analysis} {Tools} for {Machine} {Learning}},
	isbn = {978-1-4503-3145-6},
	shorttitle = {{ModelTracker}},
	url = {https://dl.acm.org/doi/10.1145/2702123.2702509},
	doi = {10.1145/2702123.2702509},
	language = {en},
	urldate = {2023-06-27},
	booktitle = {Proceedings of the 33rd {Annual} {ACM} {Conference} on {Human} {Factors} in {Computing} {Systems}},
	publisher = {ACM},
	author = {Amershi, Saleema and Chickering, Max and Drucker, Steven M. and Lee, Bongshin and Simard, Patrice and Suh, Jina},
	month = apr,
	year = {2015},
	pages = {337--346},
}

@inproceedings{sambasivan_everyone_2021,
	address = {Yokohama Japan},
	title = {“{Everyone} wants to do the model work, not the data work”: {Data} {Cascades} in {High}-{Stakes} {AI}},
	isbn = {978-1-4503-8096-6},
	shorttitle = {“{Everyone} wants to do the model work, not the data work”},
	url = {https://dl.acm.org/doi/10.1145/3411764.3445518},
	doi = {10.1145/3411764.3445518},
	language = {en},
	urldate = {2023-06-27},
	booktitle = {Proceedings of the 2021 {CHI} {Conference} on {Human} {Factors} in {Computing} {Systems}},
	publisher = {ACM},
	author = {Sambasivan, Nithya and Kapania, Shivani and Highfill, Hannah and Akrong, Diana and Paritosh, Praveen and Aroyo, Lora M},
	month = may,
	year = {2021},
	pages = {1--15},
}

@misc{klie_annotation_2022,
	title = {Annotation {Error} {Detection}: {Analyzing} the {Past} and {Present} for a {More} {Coherent} {Future}},
	shorttitle = {Annotation {Error} {Detection}},
	url = {http://arxiv.org/abs/2206.02280},
	language = {en},
	urldate = {2023-06-27},
	publisher = {arXiv},
	author = {Klie, Jan-Christoph and Webber, Bonnie and Gurevych, Iryna},
	month = sep,
	year = {2022},
	note = {arXiv:2206.02280 [cs]},
}

@article{pandey_modeling_2022,
	title = {Modeling and mitigating human annotation errors to design efficient stream processing systems with human-in-the-loop machine learning},
	volume = {160},
	issn = {10715819},
	url = {https://linkinghub.elsevier.com/retrieve/pii/S1071581922000015},
	doi = {10.1016/j.ijhcs.2022.102772},
	language = {en},
	urldate = {2023-05-17},
	journal = {International Journal of Human-Computer Studies},
	author = {Pandey, Rahul and Purohit, Hemant and Castillo, Carlos and Shalin, Valerie L.},
	month = apr,
	year = {2022},
	pages = {102772},
}

@inproceedings{gama_learning_2004,
	address = {Berlin, Heidelberg},
	series = {Lecture {Notes} in {Computer} {Science}},
	title = {Learning with {Drift} {Detection}},
	isbn = {978-3-540-28645-5},
	doi = {10.1007/978-3-540-28645-5_29},
	language = {en},
	booktitle = {Advances in {Artificial} {Intelligence} – {SBIA} 2004},
	publisher = {Springer},
	author = {Gama, João and Medas, Pedro and Castillo, Gladys and Rodrigues, Pedro},
	editor = {Bazzan, Ana L. C. and Labidi, Sofiane},
	year = {2004},
	pages = {286--295},
}

@misc{ouyang_training_2022,
	title = {Training language models to follow instructions with human feedback},
	url = {http://arxiv.org/abs/2203.02155},
	doi = {10.48550/arXiv.2203.02155},
	urldate = {2023-07-27},
	publisher = {arXiv},
	author = {Ouyang, Long and Wu, Jeff and Jiang, Xu and Almeida, Diogo and Wainwright, Carroll L. and Mishkin, Pamela and Zhang, Chong and Agarwal, Sandhini and Slama, Katarina and Ray, Alex and Schulman, John and Hilton, Jacob and Kelton, Fraser and Miller, Luke and Simens, Maddie and Askell, Amanda and Welinder, Peter and Christiano, Paul and Leike, Jan and Lowe, Ryan},
	month = mar,
	year = {2022},
	note = {arXiv:2203.02155 [cs]},
}

@inproceedings{chen_xgboost_2016,
	title = {{XGBoost}: A Scalable Tree Boosting System},
	url = {http://arxiv.org/abs/1603.02754},
	doi = {10.1145/2939672.2939785},
	shorttitle = {{XGBoost}},
	pages = {785--794},
	booktitle = {Proceedings of the 22nd {ACM} {SIGKDD} International Conference on Knowledge Discovery and Data Mining},
	author = {Chen, Tianqi and Guestrin, Carlos},
	urldate = {2023-07-27},
	date = {2016-08-13},
	eprinttype = {arxiv},
	eprint = {1603.02754 [cs]},
}

@misc{shwartz-ziv_tabular_2021,
	title = {Tabular {Data}: {Deep} {Learning} is {Not} {All} {You} {Need}},
	shorttitle = {Tabular {Data}},
	url = {http://arxiv.org/abs/2106.03253},
	doi = {10.48550/arXiv.2106.03253},
	urldate = {2023-07-27},
	publisher = {arXiv},
	author = {Shwartz-Ziv, Ravid and Armon, Amitai},
	month = nov,
	year = {2021},
	note = {arXiv:2106.03253 [cs]},
}

@article{tseng_best_2020,
	title = {Best {Practices} for {Managing} {Data} {Annotation} {Projects}},
	url = {http://arxiv.org/abs/2009.11654},
	doi = {10.13140/RG.2.2.34497.58727},
	urldate = {2023-07-27},
	author = {Tseng, Tina and Stent, Amanda and Maida, Domenic},
	year = {2020},
	note = {arXiv:2009.11654 [cs]},
}

@article{bernhardt_active_2022,
	title = {Active label cleaning for improved dataset quality under resource constraints},
	volume = {13},
	copyright = {2022 The Author(s)},
	issn = {2041-1723},
	url = {https://www.nature.com/articles/s41467-022-28818-3},
	doi = {10.1038/s41467-022-28818-3},
	language = {en},
	number = {1},
	urldate = {2023-07-27},
	journal = {Nature Communications},
	author = {Bernhardt, Mélanie and Castro, Daniel C. and Tanno, Ryutaro and Schwaighofer, Anton and Tezcan, Kerem C. and Monteiro, Miguel and Bannur, Shruthi and Lungren, Matthew P. and Nori, Aditya and Glocker, Ben and Alvarez-Valle, Javier and Oktay, Ozan},
	month = mar,
	year = {2022},
	note = {Number: 1
Publisher: Nature Publishing Group},
	pages = {1161},
}

@misc{meek_characterization_2016,
	title = {A {Characterization} of {Prediction} {Errors}},
	url = {http://arxiv.org/abs/1611.05955},
	doi = {10.48550/arXiv.1611.05955},
	urldate = {2023-07-27},
	publisher = {arXiv},
	author = {Meek, Christopher},
	month = nov,
	year = {2016},
	note = {arXiv:1611.05955 [cs]},
}

@inproceedings{rottger_two_2022,
	address = {Seattle, United States},
	title = {Two {Contrasting} {Data} {Annotation} {Paradigms} for {Subjective} {NLP} {Tasks}},
	url = {https://aclanthology.org/2022.naacl-main.13},
	doi = {10.18653/v1/2022.naacl-main.13},
	urldate = {2023-07-27},
	booktitle = {Proceedings of the 2022 {Conference} of the {North} {American} {Chapter} of the {Association} for {Computational} {Linguistics}: {Human} {Language} {Technologies}},
	publisher = {Association for Computational Linguistics},
	author = {Rottger, Paul and Vidgen, Bertie and Hovy, Dirk and Pierrehumbert, Janet},
	month = jul,
	year = {2022},
	pages = {175--190},
}

@inproceedings{hollenstein_inconsistency_2016,
	address = {Portorož, Slovenia},
	title = {Inconsistency {Detection} in {Semantic} {Annotation}},
	url = {https://aclanthology.org/L16-1629},
	urldate = {2023-07-27},
	booktitle = {Proceedings of the {Tenth} {International} {Conference} on {Language} {Resources} and {Evaluation} ({LREC}'16)},
	publisher = {European Language Resources Association (ELRA)},
	author = {Hollenstein, Nora and Schneider, Nathan and Webber, Bonnie},
	month = may,
	year = {2016},
	pages = {3986--3990},
}

@inproceedings{barnes_sentiment_2019,
	address = {Florence, Italy},
	title = {Sentiment {Analysis} {Is} {Not} {Solved}! {Assessing} and {Probing} {Sentiment} {Classification}},
	url = {https://aclanthology.org/W19-4802},
	doi = {10.18653/v1/W19-4802},
	urldate = {2023-07-27},
	booktitle = {Proceedings of the 2019 {ACL} {Workshop} {BlackboxNLP}: {Analyzing} and {Interpreting} {Neural} {Networks} for {NLP}},
	publisher = {Association for Computational Linguistics},
	author = {Barnes, Jeremy and Øvrelid, Lilja and Velldal, Erik},
	month = aug,
	year = {2019},
	pages = {12--23},
}

@inproceedings{alt_tacred_2020,
	address = {Online},
	title = {{TACRED} {Revisited}: {A} {Thorough} {Evaluation} of the {TACRED} {Relation} {Extraction} {Task}},
	shorttitle = {{TACRED} {Revisited}},
	url = {https://aclanthology.org/2020.acl-main.142},
	doi = {10.18653/v1/2020.acl-main.142},
	urldate = {2023-07-27},
	booktitle = {Proceedings of the 58th {Annual} {Meeting} of the {Association} for {Computational} {Linguistics}},
	publisher = {Association for Computational Linguistics},
	author = {Alt, Christoph and Gabryszak, Aleksandra and Hennig, Leonhard},
	month = jul,
	year = {2020},
	pages = {1558--1569},
}

@misc{northcutt_pervasive_2021,
	title = {Pervasive {Label} {Errors} in {Test} {Sets} {Destabilize} {Machine} {Learning} {Benchmarks}},
	url = {https://arxiv.org/abs/2103.14749v4},
	language = {en},
	urldate = {2023-07-27},
	journal = {arXiv.org},
	author = {Northcutt, Curtis G. and Athalye, Anish and Mueller, Jonas},
	month = mar,
	year = {2021},
}

@article{mcdonnell_why_2016,
	title = {Why {Is} {That} {Relevant}? {Collecting} {Annotator} {Rationales} for {Relevance} {Judgments}},
	volume = {4},
	copyright = {Copyright (c) 2016 Proceedings of the AAAI Conference on Human Computation and Crowdsourcing},
	issn = {2769-1349},
	shorttitle = {Why {Is} {That} {Relevant}?},
	url = {https://ojs.aaai.org/index.php/HCOMP/article/view/13287},
	doi = {10.1609/hcomp.v4i1.13287},
	language = {en},
	urldate = {2023-08-11},
	journal = {Proceedings of the AAAI Conference on Human Computation and Crowdsourcing},
	author = {McDonnell, Tyler and Lease, Matthew and Kutlu, Mucahid and Elsayed, Tamer},
	month = sep,
	year = {2016},
	pages = {139--148},
}

@inproceedings{van_halteren_detection_2000,
	address = {Centre Universitaire, Luxembourg},
	title = {The {Detection} of {Inconsistency} in {Manually} {Tagged} {Text}},
	url = {https://aclanthology.org/W00-1907},
	urldate = {2023-08-11},
	booktitle = {Proceedings of the {COLING}-2000 {Workshop} on {Linguistically} {Interpreted} {Corpora}},
	publisher = {International Committee on Computational Linguistics},
	author = {van Halteren, Hans},
	month = aug,
	year = {2000},
	pages = {48--55},
}

@inproceedings{amiri_spotting_2018,
	address = {New Orleans, Louisiana},
	title = {Spotting {Spurious} {Data} with {Neural} {Networks}},
	url = {https://aclanthology.org/N18-1182},
	doi = {10.18653/v1/N18-1182},
	urldate = {2023-08-10},
	booktitle = {Proceedings of the 2018 {Conference} of the {North} {American} {Chapter} of the {Association} for {Computational} {Linguistics}: {Human} {Language} {Technologies}, {Volume} 1 ({Long} {Papers})},
	publisher = {Association for Computational Linguistics},
	author = {Amiri, Hadi and Miller, Timothy and Savova, Guergana},
	month = jun,
	year = {2018},
	pages = {2006--2016},
}

@inproceedings{deng_imagenet_2009,
	title = {{ImageNet}: {A} large-scale hierarchical image database},
	shorttitle = {{ImageNet}},
	doi = {10.1109/CVPR.2009.5206848},
	booktitle = {2009 {IEEE} {Conference} on {Computer} {Vision} and {Pattern} {Recognition}},
	author = {Deng, Jia and Dong, Wei and Socher, Richard and Li, Li-Jia and Li, Kai and Fei-Fei, Li},
	month = jun,
	year = {2009},
	note = {ISSN: 1063-6919},
	pages = {248--255},
}

@misc{verified_market_research_data_2023,
	title = {Data {Annotation} {Service} {Market} {Size}, {Share}, {Trends} \& {Forecast}},
	url = {https://www.verifiedmarketresearch.com/product/data-annotation-service-market/},
	language = {en-US},
	urldate = {2024-05-30},
	journal = {Verified Market Research},
	author = {Verified Market Research},
	year = {2023},
}

\newpage
\section*{Supplementary Information}
\appendix
\section{XGBoost Hyperparameter Space}
\label{app:hyperparameters}

\begin{tabular*}{\textwidth}{@{\extracolsep{\fill}} lr}
\toprule
Parameter & Values \\
\midrule
classifier\_\_n\_estimators & 10, 50, 100, 150, 200, 500, 1000 \\
max\_depth & 3, 4, 5, 6, 7, 8, 9, 10 \\
min\_child\_weight & 1, 2, 3, 4, 5, 6 \\
gamma & 0, 0.1, 0.2, 0.3, 0.4, 0.5 \\
subsample & 0.6, 0.8, 1.0 \\
colsample\_bytree & 0.6, 0.7, 0.8, 0.9, 1 \\
learning\_rate & 0.01, 0.05, 0.1, 0.2, 0.3 \\
\bottomrule
\end{tabular*}

Hyperparameter grid used to train XGBoost models. Hyperparameter search was performed using randomized search.

\section{Model Performance}
\label{app:model_performance}

\begin{tabular*}{\textwidth}{@{\extracolsep{\fill}} lrrrr}
\toprule
model &       AUC &       accuracy &       precision &       recall  \\
\midrule
music streaming &  0.752345 &  0.709316 &  0.560405 &  0.678534  \\
mobile applications &  0.650713 &  0.613827 &  0.542538 &  0.607338 \\
video streaming &  0.648368 &  0.618421 &  0.538071 &  0.592672 \\
task-agnostic &  0.695857 &  0.642515 &  0.552235 &  0.643305  \\
\bottomrule
\end{tabular*}

Model performance (AUC, accuracy, precision, recall) for task-specific and task-agnostic models.

\section{Generalizability Across ML Applications}
\label{app:generalizability}

\begin{tabular*}{\textwidth}{@{\extracolsep{\fill}} lrrrr}
\toprule
{} &  music streaming &  mobile applications &  video streaming &  task-agnostic \\
\midrule
music streaming  &     0.752345 &   0.588591 &  0.613314 &      0.667225 \\
mobile applications    &     0.654686 &   0.650713 &  0.621714 &      0.649734 \\
video streaming     &     0.647696 &   0.562949 &  0.648368 &      0.610777 \\
task-agnostic &     0.744387 &   0.647421 &  0.668065 &      0.695857 \\
\bottomrule
\end{tabular*}

Generalized model performance in AUC across search modalities and product categories. Training sets are specified in rows, test sets are specified in columns.

\section{Correlations of Feature Importance Scores}
\label{app:feat_imp_cor}
\begin{tabular*}{\textwidth}{@{\extracolsep{\fill}} lrrrr}
\toprule
{} &  music streaming &  mobile applications &  video streaming &  task-agnostic \\
\midrule
music streaming  &     1.000000 &   0.375162 &  0.396875 &      0.778595 \\
mobile applications    &     0.375162 &   1.000000 &  0.261513 &      0.605633 \\
video streaming     &     0.396875 &   0.261513 &  1.000000 &      0.433407 \\
task-agnostic &     0.778595 &   0.605633 &  0.433407 &      1.000000 \\
\bottomrule
\end{tabular*}

Correlations of SHAP feature importance scores across search modalities and product categories.

\section{SHAP Feature Importance Scores}
\label{app:feat_imp_values}

\label{app:add_data}

\begin{tabular*}{\textwidth}{@{\extracolsep{\fill}} lrrrr}
\toprule
{} &  music streaming &  mobile applications &  video streaming &  task-agnostic \\
\midrule
answer\_value                          &     0.174724 &   0.018374 &  0.231036 &      0.116469 \\
comment\_length                        &     0.013941 &   0.012072 &  0.046284 &      0.044322 \\
error\_14                              &     0.073361 &   0.020695 &  0.009585 &      0.062676 \\
error\_14\_all                          &     0.015593 &   0.014081 &  0.077020 &      0.025487 \\
error\_14\_diff                         &     0.011965 &   0.032355 &  0.071587 &      0.037616 \\
error\_21                              &     0.022799 &   0.078844 &  0.018163 &      0.052727 \\
error\_21\_all                          &     0.008612 &   0.026290 &  0.017057 &      0.015521 \\
error\_21\_diff                         &     0.055123 &   0.011850 &  0.060817 &      0.020103 \\
error\_28                              &     0.219272 &   0.055531 &  0.047597 &      0.153453 \\
error\_28\_all                          &     0.018826 &   0.018202 &  0.053167 &      0.015065 \\
error\_28\_diff                         &     0.033911 &   0.026382 &  0.060519 &      0.070263 \\
error\_7                               &     0.008910 &   0.011479 &  0.011589 &      0.011565 \\
error\_7\_all                           &     0.037776 &   0.013539 &  0.026776 &      0.031598 \\
error\_7\_diff                          &     0.046419 &   0.015785 &  0.032198 &      0.019791 \\
error\_rolling\_input\_query\_type\_user\_1 &     0.000000 &   0.002750 &  0.005100 &      0.011410 \\
error\_rolling\_input\_query\_type\_user\_3 &     0.000000 &   0.003254 &  0.026068 &      0.003419 \\
error\_rolling\_input\_query\_type\_user\_5 &     0.000000 &   0.002056 &  0.017321 &      0.007626 \\
error\_rolling\_output\_media\_type\_1     &     0.007860 &   0.002154 &  0.001070 &      0.014095 \\
error\_rolling\_output\_media\_type\_3     &     0.004062 &   0.003648 &  0.005318 &      0.006995 \\
error\_rolling\_output\_media\_type\_5     &     0.000025 &   0.002037 &  0.004471 &      0.005737 \\
evaluation\_name                       &     0.000000 &   0.000000 &  0.000000 &      0.051408 \\
in\_out\_edit\_distance                  &     0.010086 &   0.029443 &  0.099182 &      0.042334 \\
in\_out\_spacy\_distance                 &     0.020118 &   0.015522 &  0.066161 &      0.047028 \\
input\_conversion\_rate                 &     0.000000 &   0.021060 &  0.037766 &      0.021376 \\
input\_language                        &     0.002164 &   0.005016 &  0.000150 &      0.023018 \\
input\_media\_type                      &     0.000000 &   0.000000 &  0.000000 &      0.000000 \\
input\_misspelled                      &     0.020800 &   0.000577 &  0.033420 &      0.001438 \\
input\_occurrences                     &     0.033666 &   0.027471 &  0.112268 &      0.046991 \\
input\_query\_type                      &     0.041456 &   0.007501 &  0.045282 &      0.101727 \\
maj\_error\_14                          &     0.008717 &   0.021240 &  0.188192 &      0.031955 \\
maj\_error\_14\_all                      &     0.063960 &   0.008968 &  0.112311 &      0.036638 \\
maj\_error\_21                          &     0.002646 &   0.007324 &  0.035188 &      0.007828 \\
maj\_error\_21\_all                      &     0.019114 &   0.010209 &  0.013003 &      0.020686 \\
maj\_error\_28                          &     0.025222 &   0.026841 &  0.108503 &      0.044418 \\
maj\_error\_28\_all                      &     0.014952 &   0.036937 &  0.045843 &      0.019300 \\
maj\_error\_7                           &     0.000141 &   0.002681 &  0.040080 &      0.019375 \\
maj\_error\_7\_all                       &     0.033265 &   0.021473 &  0.087578 &      0.028526 \\
nth\_task\_in\_session                   &     0.027264 &   0.008379 &  0.078659 &      0.018148 \\
output\_media\_type                     &     0.005976 &   0.000905 &  0.006579 &      0.010013 \\
qualification\_agreement\_rate          &     0.031739 &   0.026871 &  0.129728 &      0.038046 \\
qualification\_trials                  &     0.000956 &   0.012568 &  0.008931 &      0.009396 \\
seconds\_into\_session                  &     0.044284 &   0.018158 &  0.128082 &      0.034186 \\
storefront\_name                       &     0.044770 &   0.094815 &  0.062439 &      0.108342 \\
tenure\_full\_days                      &     0.105647 &   0.029288 &  0.131597 &      0.081711 \\
tenure\_updated\_days                   &     0.042480 &   0.039467 &  0.144892 &      0.104995 \\
time\_on\_task                          &     0.111609 &   0.036160 &  0.044017 &      0.092220 \\
vol\_last\_14                           &     0.070392 &   0.009529 &  0.011974 &      0.049065 \\
vol\_last\_21                           &     0.015969 &   0.010551 &  0.043075 &      0.042004 \\
vol\_last\_28                           &     0.011232 &   0.014959 &  0.091105 &      0.038227 \\
vol\_last\_7                            &     0.008971 &   0.018861 &  0.059923 &      0.050916 \\
\bottomrule
\end{tabular*}

Feature importance scores (cumulative SHAP values) across models. For a description of feature operationalizations, please refer to Table \ref{tab:feature_table} in the main manuscript.

\end{document}